\documentclass[10pt,twocolumn,letterpaper]{article}

\usepackage{iccv}              
\usepackage{multirow}
\usepackage[accsupp]{axessibility}

\definecolor{iccvblue}{rgb}{0.21,0.49,0.74}
\usepackage[pagebackref,breaklinks,colorlinks,allcolors=iccvblue]{hyperref}

\def\paperID{2012} 
\def\confName{ICCV}
\def\confYear{2025}

\title{All in One: Visual-Description-Guided Unified Point Cloud Segmentation}

\author{Zongyan Han$^{1}$, 
	 Mohamed El Amine Boudjoghra$^2$, 
	Jiahua Dong$^1$, \\
	Jinhong Wang$^{1}$, 
	Rao Muhammad Anwer$^{1}$
	\\ $^1$Mohamed Bin Zayed University of Artificial Intelligence, Abu Dhabi, UAE\\
    $^2$Technical University of Munich, Germany
	\\{\tt\small \{zongyan.han, jiahua.dong, jinhong.wang, rao.anwer\}@mbzuai.ac.ae}
    \\{\tt\small Mohamed.boudjoghra@tum.de}
}

\begin{document}
\maketitle
\begin{abstract}

Unified segmentation of 3D point clouds is crucial for scene understanding, but is hindered by its sparse structure, limited annotations, and the challenge of distinguishing fine-grained object classes in complex environments. Existing methods often struggle to capture rich semantic and contextual information due to limited supervision and a lack of diverse multimodal cues, leading to suboptimal differentiation of classes and instances.
To address these challenges, we propose VDG-Uni3DSeg, a novel framework that integrates pre-trained vision-language models (e.g., CLIP) and large language models (LLMs) to enhance 3D segmentation. By leveraging LLM-generated textual descriptions and reference images from the internet, our method incorporates rich multimodal cues, facilitating fine-grained class and instance separation.
We further design a Semantic-Visual Contrastive Loss to align point features with multimodal queries and a Spatial Enhanced Module to model scene-wide relationships efficiently. Operating within a closed-set paradigm that utilizes multimodal knowledge generated offline, VDG-Uni3DSeg achieves state-of-the-art results in semantic, instance, and panoptic segmentation, offering a scalable and practical solution for 3D understanding.
Our code is available at \url{https://github.com/Hanzy1996/VDG-Uni3DSeg}.

\end{abstract}    
 \section{Introduction}
Unified 3D point cloud segmentation, comprising semantic, instance, and panoptic segmentation, has become a fundamental component of tasks such as autonomous driving, robotics, and AR/VR applications.
However, progress in this field remains challenging because of the unique characteristics of point cloud data and the scarcity of annotated datasets. Unlike 2D images, which benefit from rich contextual cues such as texture and lighting, point clouds are inherently sparse and unstructured, making manual annotation labor-intensive and time-consuming.
These limitations have slowed progress in 3D segmentation, especially compared to the strides achieved in 2D segmentation, where large-scale datasets have significantly enhanced model performance. 
Although methods like OneFormer3D~\cite{kolodiazhnyi2024oneformer3d} have achieved notable success in unified 3D segmentation, they often struggle to capture fine-grained semantic details, thus limiting their effectiveness in complex 3D scenes.
 
 \begin{figure}[t]
 	\centering
 	\includegraphics[width=\linewidth]{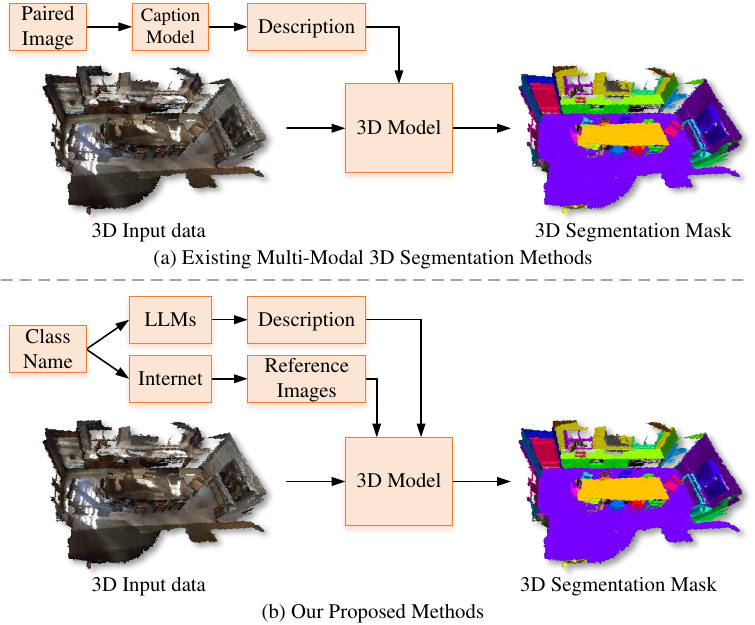}
    \vspace{-0.3cm}
 	\caption{Comparison between existing multi-modal 3D segmentation methods and our proposed approach.  
 			(a) Existing methods depend on paired images and captioning models to generate descriptions, requiring real-time 2D-3D alignment, which increases computational overhead.  
 			(b) Our approach leverages pre-generated class descriptions from LLMs and reference images retrieved from the internet, establishing offline semantic references that enhance efficiency and robustness.}
 	\vspace{-0.5cm}
 	\label{fig:intro}
 \end{figure}

To address these limitations, we propose a novel multimodal framework that leverages external knowledge from pretrained large language models (LLMs)~\cite{touvron2023llama} and vision-language models (CLIP~\cite{radford2021learning}) to integrate diverse cross-modal knowledge into the 3D segmentation pipeline.
These pretrained models aim to provide semantically rich textual and visual embeddings, enriching the segmentation process with fine-grained contextual information.
By incorporating multimodal knowledge, our approach improves the ability of the 3D segmentation model to capture fine-grained attributes crucial to distinguishing classes and instances in complex scenes, while also minimizing reliance on extensive human annotation to provide a more efficient and scalable solution for understanding 3D point clouds.
 
Specifically, our method first prompts an LLM to generate detailed semantic descriptions of 3D classes, capturing key attributes such as color, texture, and shape. These descriptions offer a broad understanding of the characteristics of the class, forming a strong semantic foundation.
To further enhance visual context, we incorporate class-specific reference images sourced from the internet, which introduce diverse real-world variations to improve the robustness of object recognition.
To effectively integrate these multi-modal cues, we encode both textual descriptions and reference images into query representations via CLIP~\cite{radford2021learning}. These queries are fused with point-wise features within a unified mask decoder, allowing the model to produce semantic, instance, and panoptic segmentation masks.
Additionally, we introduce a Semantic-Visual Contrastive (SVC) Loss, which aligns point features with their corresponding textual and visual embeddings, thereby enhancing class distinction and segmentation accuracy.
To further capture spatial relationships and refine object differentiation, we incorporate a Spatial Enhancement Module. This module utilizes sparse attention to efficiently model scene-wide spatial dependencies, ensuring structural coherence in segmentation while maintaining computational efficiency.



In particular, unlike open-vocabulary methods~\cite{ding2023pla,xiao20253d} that require real-time fusion of 3D scenes with paired images and text captions, our approach follows a closed-set paradigm with predefined class knowledge. 
We establish offline semantic references through offline-generated class descriptions and reference images (Fig.~\ref{fig:intro}).
During inference, our method operates without paired images, captioning models, or additional vision-language modules, enhancing efficiency and deployment practicality. By integrating offline multimodal knowledge with spatially aware feature learning, our approach achieves superior class separation and instance delineation compared to conventional closed-set 3D segmentation baselines.
Both quantitative and qualitative comparisons demonstrate consistent improvements in our model performance in 3D instance, semantic, and panoptic segmentation tasks.

Our contributions can be summarized as follows:
\begin{itemize}
	\item We introduce VDG-Uni3DSeg, a novel framework that integrates pretrained vision-language models and LLMs to enhance unified 3D segmentation.
	\item Our method synergizes LLM-generated descriptions and unpaired reference images, enriching class representations with both generalized attributes and fine-grained visual details, thus improving class and instance differentiation.
	\item We develop an efficient mask decoder that integrates textual and visual embeddings as queries, enhancing segmentation accuracy across 3D instance, semantic, and panoptic tasks.
\end{itemize}
\section{Related Works}
\subsection{Task specific Unimodal 3D Segmentation}
 Previous techniques for 3D segmentation adopt point-based or voxel-based approaches to process point clouds. Point-based methods use hand-crafted aggregation mechanisms~\cite{qi2017pointnet++, qian2022pointnext, thomas2019kpconv, lin2023pointmetabase} or transformer blocks~\cite{zhao2021pointtransformer, wu2022pointtransformerv2} for direct point processing, while voxel-based methods convert point clouds into sparse voxel representations, which are then processed using dense~\cite{hou20193dsis} or sparse~\cite{choy2019minkowski} 3D convolutional neural networks. Instance segmentation typically involves first performing semantic segmentation followed by per-point feature aggregation. Recent transformer-based models~\cite{schult2023mask3d} integrate sparse CNN backbones for feature extraction with transformer decoders.

The 3D instance segmentation task aims at predicting masks for individual objects in a 3D scene, along with a class label belonging to the set of known classes. Some methods use a bottom-up approach based on grouping learning embeddings in the latent space to facilitate clustering of object points~\cite{chen2021hierarchical, occuseg, he2021dyco3d, pointgroup, lahoud20193d, liang2021instance, wang2018sgpn, zhang2021point}. On the other hand, proposal-based methods adopt a top-down strategy, initially detecting 3D bounding boxes and then segmenting the object region within each box~\cite{3dmpa, hou20193d, liu2020learning, 3dbonet, yi2019gspn}. Inspired by advances in 2D work~\cite{cheng2022masked, cheng2021per}, transformer-based architectures~\cite{vaswani2017attention} have recently been introduced for 3D instance segmentation~\cite{schult2022mask3d, sun2022superpoint, kolodiazhnyi2023oneformer3d, al20233d, jain2024odin}. 
Mask3D~\cite{schult2022mask3d} introduces a hybrid architecture that uses a 3D CNN backbone for the extraction of per-point characteristics and a transformer-based instance mask decoder to refine a set of queries. 
Building on Mask3D,~\cite{al20233d} shows that explicit spatial and semantic supervision at the level of the 3D backbone further improves instance segmentation. 

For 3D panoptic segmentation,  existing methods~\cite{wu2021scenegraphfusion, narita2019panopticfusion, yang2021tuppermap} perform panoptic segmentation on RGB images, lift 2D panoptic masks into 3D space, and aggregate them to generate final 3D panoptic masks.

\subsection{Unified Unimodal segmentation}

In recent years, several approaches for 2D unified segmentation have been proposed, including MaskFormer \cite{cheng2021maskformer} and Mask2Former \cite{cheng2022mask2former}. These methods use a transformer-based architecture to predict masks for three different segmentation tasks simultaneously. OneFormer \cite{jain2023oneformer} builds on the same transformer design, but introduces task-conditioned queries, which help the model focus on a specific segmentation task during prediction.

Inspired by the success of OneFormer \cite{jain2023oneformer} in 2D, a recent method OneFormer3D \cite{kolodiazhnyi2024oneformer3d} introduces a novel multitask unified framework that simultaneously handles 3D semantic, instance, and panoptic segmentation within a single model. Unlike previous approaches, OneFormer3D unifies these tasks through semantic and instance queries in parallel within a transformer decoder, addressing the performance instability by incorporating a new query selection mechanism and an efficient matching strategy, eliminating the need for the Hungarian algorithm, which makes the training process much faster. This approach is trained once on a panoptic dataset, allowing it to outperform existing methods that are specifically tuned for each segmentation task. Although previous work has shown promising results, it still lacks essential multimodal information. In our method, we demonstrate how incorporating this information can significantly improve performance.

\section{Methodology}
\label{sec:method}
In this section, we introduce our proposed method, Visual-Description-Guided Unified 3D Segmentation (VDG-Uni3DSeg). First, we describe the 3D backbone in Sec.~\ref{sec:3D_backbone}, which extracts point-wise features from the input point cloud. Next, we detail the multi-modal reference queries in Sec.~\ref{sec:Multi-Modal Reference Queries}, which integrate textual and visual information to enhance segmentation performance. We then present the mask decoder in Sec.~\ref{sec:Multi-Modal Unified Mask Decoder}, designed to generate task-specific masks for semantic, instance, and panoptic segmentation. Finally, we describe the loss functions used to train the proposed model in Sec.~\ref{sec:Training Losses}.



\subsection{3D Backbone}\label{sec:3D_backbone} 
The input to our model is a 3D scene represented as a point cloud $\mathbf{P} \in \mathbb{R}^{N \times 6}$ containing $N$ points. Each point consists of three spatial coordinates and three color values. To expedite the training process and improve computational efficiency, we adopt the voxelization approach from~\cite{choy20194d}, which reduces the point cloud to $M$ superpoints with $M < N$, allowing the model to focus on meaningful spatial clusters.  
Subsequently, these superpoints are processed by a sparse 3D U-Net to generate point-wise features, $\mathbf{X} \in \mathbb{R}^{M \times d}$, where $d$ denotes the point embedding dimensionality.

\begin{figure*}[ht!]
	\centering
	\includegraphics[width=\linewidth]{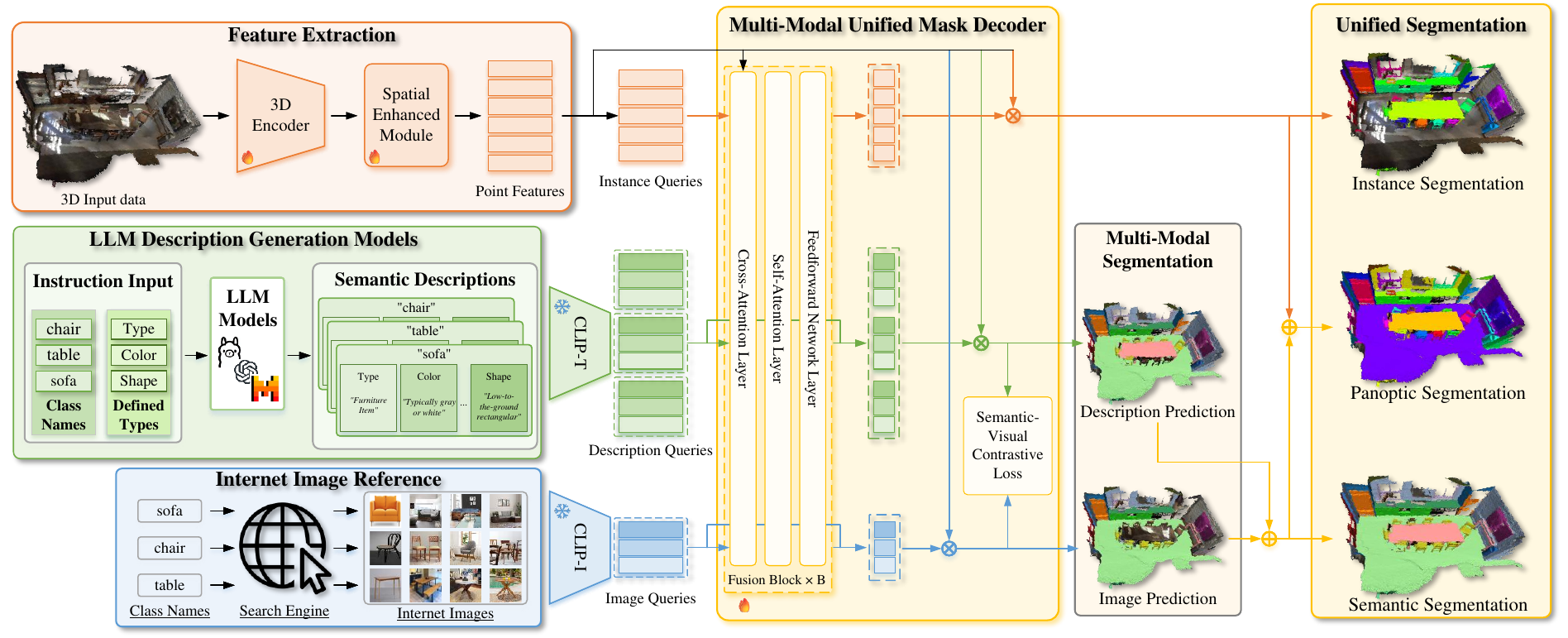}
	\caption{Illustration of the \textbf{VDG-Uni3DSeg} framework.  
		First, we utilize a 3D encoder to extract point-wise features from the input point cloud.  
		Then, the Spatial Enhancement Module refines these features by capturing local spatial relationships.  
		Next, class-specific description queries are generated using LLMs, while image queries are constructed from reference images retrieved from the internet. Both description and image queries are embedded using CLIP. These queries, along with the enhanced point features, are processed through a multi-modal fusion block to generate task-specific masks.  
		The description-based and image-based predictions are then fused to achieve robust semantic segmentation.  
		Finally, instance and semantic segmentation masks are integrated to produce panoptic segmentation results.
	}
	\label{fig:main}
\end{figure*}

\paragraph{Spatial Enhancement Module}
To capture the spatial relationships between point-wise features in a 3D scene, we adopt a sparse attention mechanism~\cite{child2019generating} that integrates contextual information from a randomly sampled subset of the entire point cloud.
Specifically, the point-wise features $\mathbf{X}$ are first transformed into query, key, and value embeddings:
\begin{equation} 
	\mathbf{Q} = \mathbf{X} \mathbf{W}_q, \ \mathbf{K} = \mathbf{X} \mathbf{W}_k, \ \mathbf{V} = \mathbf{X} \mathbf{W}_v, 
\end{equation}
where $\mathbf{W}_q$, $\mathbf{W}_k$, and $\mathbf{W}_v \in \mathbb{R}^{d \times d}$ are learnable weight matrices.
For each point feature $\mathbf{x}_i$, attention is calculated using a randomly sampled subset $\mathcal{S}_i$. 
The spatially enhanced feature  is then computed as:
\begin{equation} 
		\widetilde{\mathbf{x}}_i = \sum_{\mathbf{v}_j \in \mathcal{S}_i} \alpha_{i, j} \cdot \mathbf{v}_j,  
		\ \ \ 
		\alpha_{i, j}=\frac{\exp (\mathbf{q}_i \cdot \mathbf{k}_{j}^{\top}/{\sqrt{d}})}{\sum_{{\mathbf{k}_j} \in \mathcal{S}_i} \exp (\mathbf{q}_i \cdot \mathbf{k}_{j}^{\top}/{\sqrt{d}})},
\end{equation}
where $\mathcal{S}_i$ denotes a randomly sampled subset of points from the point cloud, and $\mathbf{q}_i$, $\mathbf{k}_j$, $\mathbf{v}_j \in \mathbb{R}^{1 \times d}$. 
Using the subset $\mathcal{S}_i$, the module captures the spatial structure of the overall scene while significantly reducing computational costs. This makes it highly efficient and well-suited for large-scale 3D point cloud processing.

\subsection{Multi-Modal Reference Queries}\label{sec:Multi-Modal Reference Queries}
Our approach uses a unified 3D mask decoder to perform semantic, instance, and panoptic segmentation simultaneously. 
Specifically, we build on the Transformer-based mask decoder framework proposed in~\cite{jain2023oneformer}. For each segmentation task, we incorporate task-specific queries that interact with the point-wise features to produce the corresponding masks.
For instance segmentation, we use instance queries as defined in~\cite{jain2023oneformer}, denoted as $\mathbf{Q}_{ins} \in \mathbb{R}^{K \times d}$.
For semantic segmentation, we introduce multimodal semantic queries to generate masks as described below. 

\noindent \textbf{Description Reference Queries}
To capture fine-grained details for accurate class distinction, we first introduce the LLM-based Description Generator (LLM-DesGen), a pipeline designed to generate descriptive visual details specific to the class.
LLM-DesGen is guided by pre-defined description types and example visual descriptions for given classes.
In this in-context learning manner, it can be generalized to unseen classes by learning the descriptive patterns. 
For each candidate class, LLM-DesGen generates $K$ descriptions for the given class, as follows:
\begin{equation}
	\mathbf{T}^c=[t^c_1, \dots, t^c_K] = \text{LLM-DesGen}(c),
\end{equation}
where $\mathbf{T}^c$ represents the descriptions of the $c$-th class,  and $t^c_k$ denotes the $k$-th description of the $c$-th class.
Finally, we can get the set of descriptions for all classes: $\mathbf{T} = [\mathbf{T}^1, \dots, \mathbf{T}^C]$.
Next, we utilize a text encoder (i.e., CLIP~\cite{radford2021learning}) to embed these descriptions into description embeddings:
\begin{equation}
	\mathbf{Q}_{t} = \text{CLIP-T}(\mathbf{T}), 
\end{equation}
where $\mathbf{Q}_{t} = [\mathbf{Q}_{t}^{1}, \dots, \mathbf{Q}_{t}^{C}]$, where $\mathbf{Q}_{t}^{c} \in \mathbb{R}^{K \times d}$ denotes the $K$ queries for class $c$

\noindent \textbf{Image Reference Queries}
To complement the class descriptions, we collect a set of reference images for each class by collecting images from the internet, denoted $\mathbf{O} =[\mathbf{O}^1, \dots, \mathbf{O}^C]$, where $\mathbf{O}^c = [o_1^c, \dots, o_L^c]$ consists of $L$ images representing the $c$-th class. We embed these images using the CLIP image encoder~\cite{radford2021learning} to generate visual embeddings as follows:
\begin{equation}
	\mathbf{Q}_{o} = \text{CLIP-I}(\mathbf{O}), 
\end{equation}
where $\mathbf{Q}_{o} \in \mathbb{R}^{C \times L \times d}$. Unlike multiview images in standard training datasets, these internet-sourced images provide diverse representations with broader variations, reducing reliance on extensive annotations. 
 
In Uni3DSeg, description queries provide high-level semantic information, capturing the general appearance and characteristics of each class, while visual queries offer fine-grained visual details.

\subsection{Multi-Modal Unified Mask Decoder}\label{sec:Multi-Modal Unified Mask Decoder}
Following the approach in~\cite{kolodiazhnyi2024oneformer3d}, we first utilize $B$ fusion layers to align and blend the information from the point cloud, image, and description domains, enhancing the model's capacity for accurate multimodal mask prediction. Concretely, we apply cross-attention to integrate 3D information into our task-specific queries. Then, we use a self-attention mechanism and a feed-forward network to combine the cross-modal information within these queries.
To preserve the original semantic and visual information, we incorporate a residual structure into the description and image reference queries, as shown in Fig.~\ref{fig:main}. 
We then feed these refined queries and the spatially enhanced point features $\widetilde{\mathbf{X}} $ into the main mask decoder $D$ to generate the corresponding segmentation masks for each query:
\begin{small}
	\begin{align}
		\!\!\!\mathbf{M}_{ins} = D(\mathbf{Q}_{ins}, \widetilde{\mathbf{X}}), 
		\mathbf{M}_{t} = D(\mathbf{Q}_{t}, \widetilde{\mathbf{X}}), 
		\mathbf{M}_{o} = D(\mathbf{Q}_{o}, \widetilde{\mathbf{X}}),\!\!\!
		\label{eq: vd}
	\end{align}
\end{small}where $\mathbf{M}_{ins}\in \mathbb{R}^{ N\times S}$ is the predicted instance segmentation mask, $\mathbf{M}_{t}\in \mathbb{R}^{N\times C \times K} $ and $\mathbf{M}_{o}\in \mathbb{R}^{N\times C \times L} $ are the semantic segmentation masks predicted by the description and image queries, respectively.

\noindent \textbf{Multi-Modal Segmentation Ensemble}
To achieve a robust semantic segmentation result, we adopt an ensemble strategy that combines the semantic segmentation outputs of $\mathbf{M}_{d}$ and $\mathbf{M}_{o}$.
Concretely, for each point feature $\mathbf{x}_m$, the description-based semantic prediction  $\mathbf{M}_{d}^{m}  \in \mathbb{R}^{C \times K}$ represents the similarity scores for all descriptions for each class.
To determine the best-matching description for each point, we apply max pooling on $\mathbf{M}_{d}^{m}$ to obtain the refined description-based prediction $\widehat{\mathbf{M}}_{d}^{m} \in \mathbb{R}^{C}$:
\begin{equation}
	\widehat{\mathbf{M}}_{t}^{m}  = \max_k (\mathbf{M}_{t}^{m}).
\end{equation}
Similarly, we obtain the refined image prediction $\widehat{\mathbf{M}}_{o}^{m}$. By combining description-based and image-based predictions, the ensemble approach produces a robust semantic segmentation result that leverages both high-level semantic and fine-grained visual details. 
The final semantic segmentation prediction is computed as:
\begin{equation}
	{\mathbf{M}}_{sem}^{m}  = 	\widehat{\mathbf{M}}_{t}^{m}  + 	\widehat{\mathbf{M}}_{o}^{m},
\end{equation}
where ${\mathbf{M}}_{sem}^{m}  \in \mathbb{R}^{C}$ is the semantic prediction of the $m$-th point feature. 
The final result of semantic segmentation is denoted as ${\mathbf{M}}_{sem}  \in \mathbb{R}^{M\times C}$.

\subsection{Training Losses}\label{sec:Training Losses}

Following \cite{jain2023oneformer}, we use a combination of cross-entropy, binary cross-entropy, and Dice losses to optimize the instance mask predictions.
For semantic segmentation, in addition to cross-entropy loss, we introduce the Semantic-Visual Contrastive (SVC) Loss, which aligns point features with both description embeddings $\mathbf{Q}_{d}$ and image embeddings $\mathbf{Q}_{o}$. 
In contrast to traditional contrastive losses that only align features with a single-modality embedding, the SVC loss promotes multimodal alignment, encouraging spatially enhanced point features $\widetilde{\mathbf{X}}$ to be close to both description embeddings $\mathbf{Q}_{d}$ and image embeddings $\mathbf{Q}_{o}$ of the correct class while pushing them farther from the embeddings of other classes. The SVC loss is defined as follows:
\begin{small}
	\begin{equation}
		\begin{aligned}
			&\mathcal{L}_{\text {SVC}}= \mathcal{L}_{\text {c}}(\widetilde{\mathbf{X}}, \mathbf{Q}_{t}) +	\mathcal{L}_{\text {c}}(\widetilde{\mathbf{X}}, \mathbf{Q}_{o}),\\
			&\mathcal{L}_{\text {c}}(\widetilde{\mathbf{X}}, \mathbf{Q})= -\mathbb{E}_{(\tilde{\mathbf{x}}_i, \mathbf{q}^+)} \left[\log \frac{\exp (\operatorname{sim}(\tilde{\mathbf{x}}_i, \mathbf{q}^+) / \tau)}{\sum_{n} \exp (\operatorname{sim}(\tilde{\mathbf{x}}_i, \mathbf{q}_n) / \tau)}\right],
		\end{aligned}
	\end{equation}
\end{small}where $\mathbf{q}_+$ denotes the positive description or image embedding corresponding to the same class as the point feature $\tilde{\mathbf{x}}_i$,  $\operatorname{sim}(\cdot)$ denotes the inner product similarity, and $\tau$ is the temperature parameter, empirically set to 1.0.
This loss aligns point features with class-specific embeddings, enhancing both semantic consistency and class discriminability. 
It also facilitates better integration of information from reference descriptions and images, contributing to a more robust segmentation.

The overall loss function for our method is defined as:
\begin{equation}
	\mathcal{L}_{all}=  \underbrace{\mathcal{L}_{bce} + \mathcal{L}_{dice}+ \lambda_1\mathcal{L}_{ce} ^ {ins}}_{\mathcal{L}_{ins}} +	 \underbrace{\lambda_2\mathcal{L}^{sem}_{ce} +\lambda_3 \mathcal{L}_{SVC}}_{\mathcal{L}_{sem}},
\end{equation}where $\lambda_1$, $\lambda_2$, and $\lambda_3$ are hyperparameters that balance different loss functions.
\section{Experiments}
\subsection{Experimental Setting}
\paragraph{Datasets}
We evaluate our method on three widely used datasets: S3DIS~\cite{armeni20163d}, ScanNet~\cite{dai2017scannet}, and ScanNet200~\cite{rozenberszki2022language}.
The S3DIS dataset~\cite{armeni20163d} comprises 272 scenes spread over six areas, with annotations for 13 semantic classes. 
Among these, five furniture categories are defined as thing classes, while the remaining eight are stuff classes for panoptic evaluation. Following the standard evaluation protocol, we evaluate our method on Area-5 and conduct six-fold cross-validation.
ScanNet dataset~\cite{dai2017scannet} consists of 1,201 scans for training, 312 scans for validation, and 100 scans for testing. The 3D instance segmentation task includes 18 object classes, while the semantic and panoptic segmentation tasks incorporate two additional background classes.
ScanNet200 dataset~\cite{rozenberszki2022language} extends ScanNet~\cite{dai2017scannet} with finer-grained annotations, containing 198 object classes for instance segmentation and two additional background classes. 
It follows the same data split as ScanNet~\cite{dai2017scannet}.

\begin{table*}[ht!]
	\centering
	\small
	  \resizebox{0.7\linewidth}{!}
	{
	\begin{tabular}{llcccccccc}
		\toprule
		& \multirow{2}{*}{Method} & \multicolumn{4}{c}{Instance} & Semantic & \multicolumn{3}{c}{Panoptic} \\
		& & mAP\textsubscript{50} & mAP & mPrec\textsubscript{50} & mRec\textsubscript{50} & mIoU & PQ & PQ\textsubscript{th} & PQ\textsubscript{st} \\
		\midrule
		\multicolumn{2}{l}{\textit{Area-5 validation}} & & & & & & & & \\
		& PointGroup\cite{jiang2020pointgroup} & 57.8 & & 61.9 & 62.1 & & & & \\
		& DyCo3D\cite{he2021dyco3d} & & & 64.3 & 64.2 & & & & \\
		& SSTNet\cite{liang2021sstnet} & 59.3 & 42.7 & 65.5 & 64.2 & & & & \\
		& DKNet\cite{wu2022dknet} & & & 70.8 & 65.3 & & & & \\
		& HAIS\cite{chen2021hais} & & & 71.1 & 65.0 & & & & \\
		& TD3D\cite{kolodiazhnyi2023td3d} & 65.1 & 48.6 & 74.4 & 64.8 & & & & \\
		& SoftGroup\cite{vu2022softgroup} & 66.1 & 51.6 & 73.6 & 66.6 & & & & \\
		& PBNet\cite{zhao2023pbnet} & 66.4 & 53.5 & 74.9 & 65.4 & & & & \\
		& SPFormer\cite{sun2023spformer} & 66.8 & & 72.8 & 67.1 & & & & \\
		& Mask3D\cite{schult2023mask3d} & 71.9 & 57.8 & 74.3 & 63.7 & & & & \\
		& SegGCN\cite{lei2020seggcn} & & & & & 63.6 & & & \\
		& MinkUNet\cite{choy2019minkowski} & & & & & 65.4 & & & \\
		& PAConv\cite{xu2021paconv} & & & & & 66.6 & & & \\
		& KPConv\cite{thomas2019kpconv} & & & & & 67.1 & & & \\
		& PointTransformer\cite{zhao2021pointtransformer} & & & & & 70.4 & & & \\
		& PointNeXt-XL\cite{qian2022pointnext} & & & & & 70.5 & & & \\
		& PointTransformerV2\cite{wu2022pointtransformerv2} & & & & & \textbf{71.6} & & & \\
		& OneFormer3D~\cite{kolodiazhnyi2024oneformer3d} & {72.0} & {58.7} & {79.7} & {73.0} & 69.8 & {62.2} & {58.4} & \textbf{65.5} \\
		&\textbf{VDG-Uni3DSeg (Ours)} & \textbf{74.1} & \textbf{60.1} & \textbf{81.0} & \textbf{73.9} & {71.5} & \textbf{66.3} &\textbf{68.0} & {65.3}  \\
		\midrule
		\multicolumn{2}{l}{\textit{6-fold cross-validation}} & & & & & & & & \\
		& PointGroup\cite{jiang2020pointgroup} & 64.0 & & 69.6 & 69.2 & & & & \\
		& HAIS\cite{chen2021hais} & & & 73.2 & 69.4 & & & & \\
		& SSTNet\cite{liang2021sstnet} & 67.8 & 54.1 & 73.5 & 73.4 & & & & \\
		& DKNet\cite{wu2022dknet} & & & 75.3 & 71.1 & & & & \\
		& TD3D\cite{kolodiazhnyi2023td3d} & 68.2 & 56.2 & 76.3 & 74.0 & & & & \\
		& SoftGroup\cite{vu2022softgroup} & 68.9 & 54.4 & 75.3 & 69.8 & & & & \\
		& SPFormer\cite{sun2023spformer} & 69.2 & & 74.0 & 71.1 & & & & \\
		& PBNet\cite{zhao2023pbnet} & 70.6 & 59.5 & 80.1 & 72.9 & & & & \\
		& Mask3D\cite{schult2023mask3d} & 74.3 & 61.8 & 76.5 & 66.2 & & & & \\
		& PointNet++\cite{qi2017pointnet++} & & & & & 56.7 & & & \\
		& MinkUNet\cite{choy2019minkowski} & & & & & 69.1 & & & \\
		& KPConv\cite{thomas2019kpconv} & & & & & 70.6 & & & \\
		& PointTransformer\cite{zhao2021pointtransformer} & & & & & 73.5 & & & \\
		& PointNeXt-XL\cite{qian2022pointnext} & & & & &\textbf{ 74.9} & & & \\
		& OneFormer3D~\cite{kolodiazhnyi2024oneformer3d} & {72.8} & {60.4} & {82.3} & {72.7} & {72.8} & {68.5} & {61.5} & {\textbf{74.5}} \\
		& \textbf{VDG-Uni3DSeg (Ours)} &\textbf{75.8}&\textbf{63.7}&\textbf{85.5}&\textbf{74.8}&73.2&\textbf{69.1}&\textbf{68.9}&72.7\\
		
		\bottomrule
	\end{tabular}
}
	\caption{Comparison of existing segmentation methods on S3DIS. Our proposed VDG-Uni3DSeg shows significant performance improvement across instance, semantic, and panoptic segmentation tasks.}
	\label{tab:s3dis}
\end{table*}

\noindent \textbf{Evaluation Metrics}
For 3D semantic segmentation, we evaluate performance based on the mean Intersection over Union (mIoU). 
For 3D instance segmentation, we compute the mean Average Precision (mAP) across IoU thresholds ranging from 50\% to 95\%, with increments of 5\%. 
Furthermore, mAP50 and mAP25 represent mAP values calculated at IoU thresholds of 50\% and 25\%, respectively.
For S3DIS, we additionally report mean precision (mPrec) and mean recall (mRec) according to the standard evaluation protocol. 
3D panoptic segmentation is evaluated using the Panoptic Quality (PQ) score~\cite{kirillov2019panoptic}. 
We also provide separate evaluations for the categories of things and stuff, denoted PQ\textsubscript{th} and PQ\textsubscript{st}, respectively.

\noindent \textbf{Implementation Details}
We implement our method using the MMDetection3D framework~\cite{mmdet3d2020}, following OneFormer3D~\cite{kolodiazhnyi2024oneformer3d}.
The training setup utilizes AdamW optimizer with an initial learning rate of 0.0001, a weight decay of 0.05, a batch size of 4, and a polynomial learning rate scheduler with a power of 0.9.
We apply standard data augmentations, including horizontal flipping, random rotations along the z-axis, elastic distortion, and random scaling.
For each class, we generate 10 descriptions using LLaMA 3.1~\cite{touvron2023llama} and collect reference images from Bing.
We then extract the description and image embeddings using CLIP~\cite{radford2021learning}.
These embeddings are further projected into the 3D feature space as queries via a multi-layer perceptron.


\begin{table*}[ht!]
	\centering
	\small
		  \resizebox{0.75\linewidth}{!}
	{
	\begin{tabular}{lllccccccc}
		\toprule
		& \multirow{2}{*}{Method} & \multirow{2}{*}{Presented at} & \multicolumn{3}{c}{Instance} & Semantic & \multicolumn{3}{c}{Panoptic} \\
		& & & mAP\textsubscript{25} & mAP\textsubscript{50} & mAP & mIoU & PQ & PQ\textsubscript{th} & PQ\textsubscript{st} \\
		\midrule
		\multicolumn{2}{l}{\textit{Validation split}} & & & & & & & & \\
		& NeuralBF\cite{sun2023neuralbf} & WACV'23 & 71.1 & 55.5 & 36.0 & & & & \\
		& DyCo3D\cite{he2021dyco3d} & CVPR'21 & 72.9 & 57.6 & 35.4 & & & & \\
		& SSTNet\cite{liang2021sstnet} & ICCV'21 & 74.0 & 64.3 & 49.4 & & & & \\
		& HAIS\cite{chen2021hais} & ICCV'21 & 75.6 & 64.4 & 43.5 & & & & \\
		& DKNet\cite{wu2022dknet} & ICCV'22 & 76.9 & 66.7 & 50.8 & & & & \\
		& SoftGroup\cite{vu2022softgroup} & CVPR'22 & 78.9 & 67.6 & 45.8 & & & & \\
		& PBNet\cite{zhao2023pbnet} & ICCV'23 & 78.9 & 70.5 & 54.3 & & & & \\
		& TD3D\cite{kolodiazhnyi2023td3d} & WACV'24 & 81.9 & 71.2 & 47.3 & & & & \\
		& ISBNet\cite{ngo2023isbnet} & CVPR'23 & 82.5 & 73.1 & 54.5 & & & & \\
		& SPFormer\cite{sun2023spformer} & AAAI'23 & 82.9 & 73.9 & 56.3 & & & & \\
		& Mask3D\cite{schult2023mask3d} & ICRA'23 & 83.5 & 73.7 & 55.2 & & & & \\
		& PointTransformer\cite{zhao2021pointtransformer} & ICCV'21 & & & & 70.6 & & & \\
		& PointNeXt-XL\cite{qian2022pointnext} & NeurIPS'22 & & & & 71.5 & & & \\
		& PointMetaBase-XXL\cite{lin2023pointmetabase} & CVPR'23 & & & & 72.8 & & & \\
		& PointTransformerV2\cite{wu2022pointtransformerv2} & NeurIPS'22 & & & & 75.4 & & & \\
		& SceneGraphFusion\cite{wu2021scenegraphfusion} & CVPR'21 & & & & & 31.5 & 30.2 & 43.4 \\
		& PanopticFusion\cite{narita2019panopticfusion} & IROS'19 & & & & & 33.5 & 30.8 & 58.4 \\
		& TUPPer-Map\cite{yang2021tuppermap} & IROS'21 & & & & & 50.2 & 47.8 & 71.5 \\
		& {OneFormer3D} ~\cite{kolodiazhnyi2024oneformer3d}&CVPR'24 & 86.1&77.2&58.5&75.8&71.3&\textbf{70.5}&86.1\\
		
		&  \textbf{VDG-Uni3DSeg (Ours)}  & & \textbf{86.5}& \textbf{78.5}& \textbf{59.3}& \textbf{76.2}& \textbf{71.5}&{70.0}&\textbf{86.6}\\
		
		\midrule
		\multicolumn{2}{l}{\textit{Hidden test split at {03 Mar. 2025}}} & & & & & & & & \\
		& NeuralBF\cite{sun2023neuralbf} & WACV'23 & 71.8 & 55.5 & 35.3 & & & & \\
		& DyCo3D\cite{he2021dyco3d} & CVPR'21 & 76.1 & 64.1 & 39.5 & & & & \\
		& PointGroup\cite{jiang2020pointgroup} & CVPR'20 & 77.8 & 63.6 & 40.7 & & & & \\
		& SSTNet\cite{liang2021sstnet} & ICCV'21 & 78.9 & 69.8 & 50.6 & & & & \\
		& HAIS\cite{chen2021hais} & ICCV'21 & 80.3 & 69.9 & 45.7 & & & & \\
		& DKNet\cite{wu2022dknet} & ICCV'22 & 81.5 & 71.8 & 53.2 & & & & \\
		& ISBNet\cite{ngo2023isbnet} & CVPR'23 & 83.5 & 75.7 & 55.9 & & & & \\
		& SPFormer\cite{sun2023spformer} & AAAI'23 & 85.1 & 77.0 & 54.9 & & & & \\
		& SoftGroup\cite{vu2022softgroup} & CVPR'22 & 86.5 & 76.1 & 50.4 & & & & \\
		& Mask3D\cite{schult2023mask3d} & ICRA'23 & 87.0 & 78.0 & 56.6 & & & & \\
		& TD3D\cite{kolodiazhnyi2023td3d} & WACV'24 & 87.5 & 75.1 & 48.9 & & & & \\
		& OneFormer3D~\cite{kolodiazhnyi2024oneformer3d} & CVPR'24& \textbf{89.6} & {80.1} & {56.6} & & & & \\
		&  \textbf{VDG-Uni3DSeg (Ours)}   &  &88.0 & \textbf{80.4 }& \textbf{57.6}&  & &  & \\
		\bottomrule
	\end{tabular}
}
	\caption{Comparison of existing segmentation methods on ScanNet. Our proposed VDG-Uni3DSeg achieves superior performance across all metrics for instance, semantic, and panoptic segmentation tasks. 
    }
	\label{tab:scannet}
\end{table*}

\subsection{Comparisons with SOTA Methods}
We compare our method against state-of-the-art (SOTA) approaches on three datasets.
Table~\ref{tab:s3dis} presents comparisons on the S3DIS~\cite{armeni20163d} dataset.
Across both the Area-5 and 6-fold cross-validation benchmarks, our method demonstrates strong competitive performance compared to SOTA methods.
Notably, on the Area-5 benchmark, our method achieves significant improvements in instance segmentation, outperforming previous methods by a large margin.
Specifically, VDG-Uni3DSeg achieves a 2.1-point gain in mAP\textsubscript{50} and a 1.4-point improvement in mAP over the best-performing baseline. 
For panoptic segmentation, our method surpasses SOTA methods with a 4.1-point gain in PQ and a remarkable 9.6-point improvement in PQ\textsubscript{th}.
Additionally, for semantic segmentation, VDG-Uni3DSeg achieves a 1.7-point improvement in mIoU over our baseline, OneFormer3D~\cite{kolodiazhnyi2024oneformer3d}.
On the 6-fold cross-validation benchmark, VDG-Uni3DSeg achieves similarly notable improvements across all metrics, demonstrating its robustness and effectiveness in different settings.

\begin{table*}[ht!]
	\centering
	\small
			  \resizebox{0.65\linewidth}{!}
	{
	\begin{tabular}{lccccccc}
		\toprule
		\multirow{2}{*}{Method} & \multicolumn{3}{c}{Instance} & Semantic & \multicolumn{3}{c}{Panoptic} \\
		& mAP\textsubscript{25} & mAP\textsubscript{50} & mAP & mIoU & PQ & PQ\textsubscript{th} & PQ\textsubscript{st}\\
		\midrule
		PointGroup\cite{jiang2020pointgroup} & & 24.5 & & & & & \\
		PointGroup + LGround\cite{rozenberszki2022lground} & & 26.1 & & & & & \\
		TD3D\cite{kolodiazhnyi2023td3d} & 40.4 & 34.8 & 23.1 & & & & \\
		Mask3D\cite{schult2023mask3d} & 42.3 & 37.0 & 27.4 & & & & \\
		MinkUNet\cite{choy2019minkowski} & & & & 25.0 & & & \\
		MinkUNet + LGround\cite{rozenberszki2022lground} & & & & 28.9 & & & \\
	   {OneFormer3D}~\cite{kolodiazhnyi2024oneformer3d} & 44.2&40.0&29.0&29.0&31.0&30.5&\textbf{78.6}\\
    	\textbf{VDG-Uni3DSeg (Ours)}  &  \textbf{45.1}&\textbf{40.0} & \textbf{29.5}& \textbf{29.7}& \textbf{31.3}& \textbf{30.9}&77.9 \\
		\bottomrule
	\end{tabular}
}
	\caption{Comparison of existing segmentation methods on the ScanNet200 validation split. Our VDG-Uni3DSeg shows competitive performance across instance, semantic, and panoptic segmentation tasks.}
    \vspace{-0.3cm}
	\label{tab:scannet200}
\end{table*}

\begin{table*}[ht!]
	\centering
	\small
	 \resizebox{0.65\linewidth}{!}
	{
	\begin{tabular}{lccccccc}
		\toprule
		\multirow{2}{*}{Method} & \multicolumn{3}{c}{Instance} & Semantic & \multicolumn{3}{c}{Panoptic} \\
		& mAP\textsubscript{25} & mAP\textsubscript{50} & mAP & mIoU & PQ & PQ\textsubscript{th} & PQ\textsubscript{st}\\
		\midrule
		baseline &79.5& {72.0} & {58.7} &69.8 & {62.2} & {58.4} & {65.5} \\
         \textbf{VDG-Uni3DSeg} & \textbf{80.6} & \textbf{74.1} & \textbf{60.1} & \textbf{71.5} & \textbf{66.3} & {68.0} & {65.3}  \\
		 \midrule
		W/O Spatial Enhancement&80.0&73.3&58.1&70.6&64.6&63.8&\textbf{65.7}\\
	 W/O Description Query &78.4&72.4&59.6&70.2&64.8&\textbf{68.9}&62.2\\
	W/O Image Query  & 79.7 & 72.7& 60.0& 70.4& 65.1& 64.3& {65.6}\\
	W/O SVC Loss&78.0& 72.5& 58.3& 69.0& 62.2&61.1& 63.8\\
		\bottomrule
	\end{tabular}
}
	\caption{Ablation study evaluating each component of our VDG-Uni3DSeg. `W/O' indicates that the corresponding part of our VDG-Uni3DSeg is removed. 
    }
   \vspace{-0.1cm}
	\label{tab:ablation}
\end{table*}

Table~\ref{tab:scannet} presents comparisons on the ScanNet~\cite{dai2017scannet} dataset, including results for both the validation split and the hidden test split.
In the validation split, our method consistently improves performance across all segmentation tasks, including instance, semantic, and panoptic segmentation.
Specifically, for instance segmentation, VDG-Uni3DSeg achieves a 1.3-point improvement in mAP\textsubscript{50}, demonstrating its enhanced ability to detect and distinguish individual objects.
For semantic segmentation, our method achieves a 0.4-point improvement in mIoU, highlighting its improved capability to classify points accurately across the scene.
Furthermore, for panoptic segmentation, VDG-Uni3DSeg also achieves improvement in PQ, demonstrating its effectiveness in jointly optimizing instance and semantic predictions.
Additionally, VDG-Uni3DSeg also achieves improvements in the ScanNet hidden test set, further highlighting its strong generalization ability.
Table~\ref{tab:scannet200} presents the results for ScanNet200~\cite{rozenberszki2022language}, which introduces fine-grained classes.
VDG-Uni3DSeg achieves state-of-the-art performance in all instance, semantic, and panoptic segmentation, demonstrating its robustness in complex segmentation tasks.
Although our model achieves slightly lower semantic mIoU, it remains competitive and maintains balanced performance across all tasks.

These consistent improvements across different datasets highlight the effectiveness of using both visual and description-based references to guide the segmentation process. By integrating multi-modal information, our method enhances the understanding of objects in 3D scenes, leading to significant performance gains.

\begin{figure*}[ht!]
	\centering
	\begin{subfigure}[b]{0.19\textwidth}
		\centering
		\includegraphics[width=\textwidth]{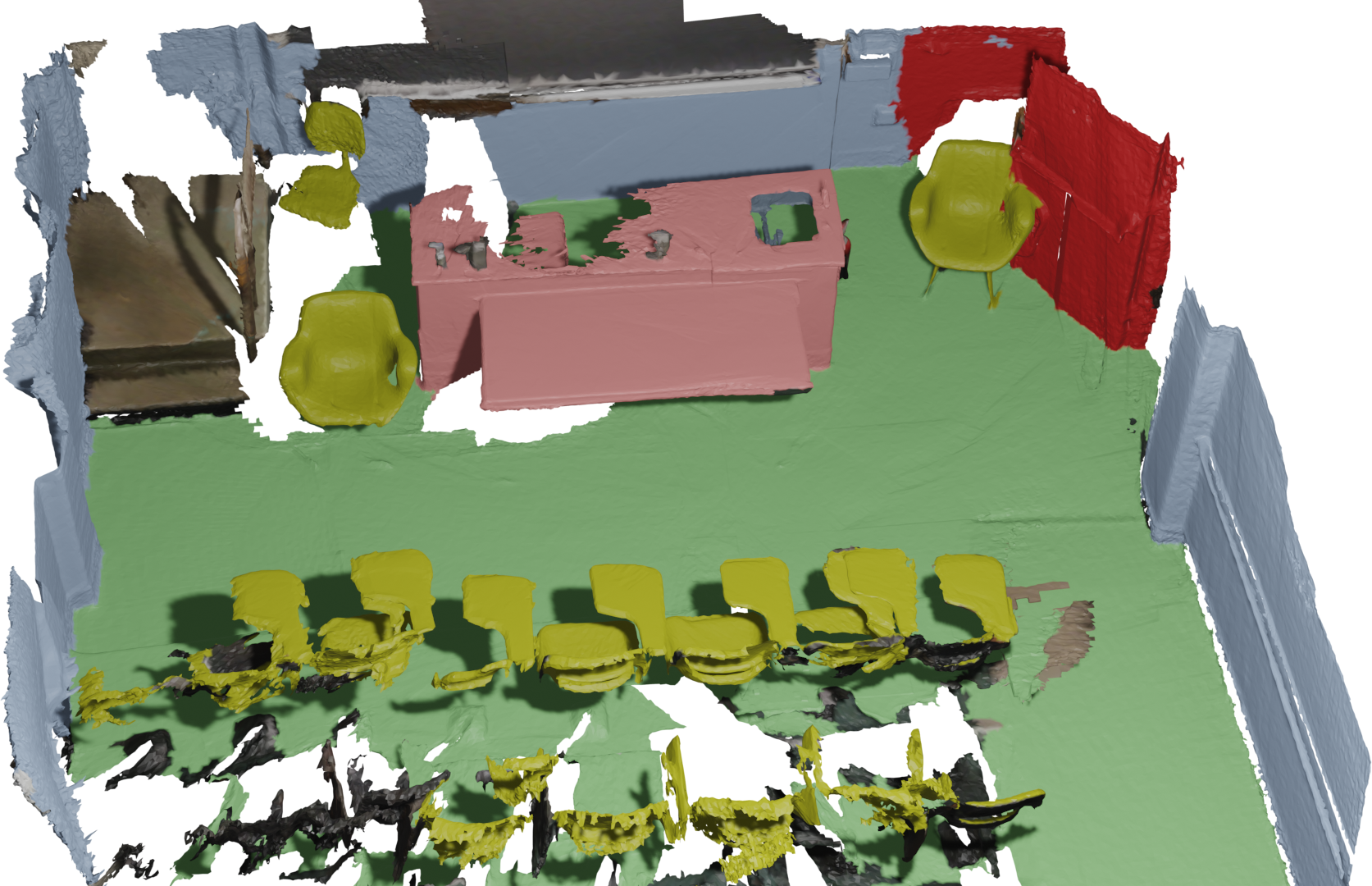}
		\caption{\footnotesize{Ground Truth}}
		\label{fig:gt}
	\end{subfigure}
	\begin{subfigure}[b]{0.19\textwidth}
		\centering
		\includegraphics[width=\textwidth]{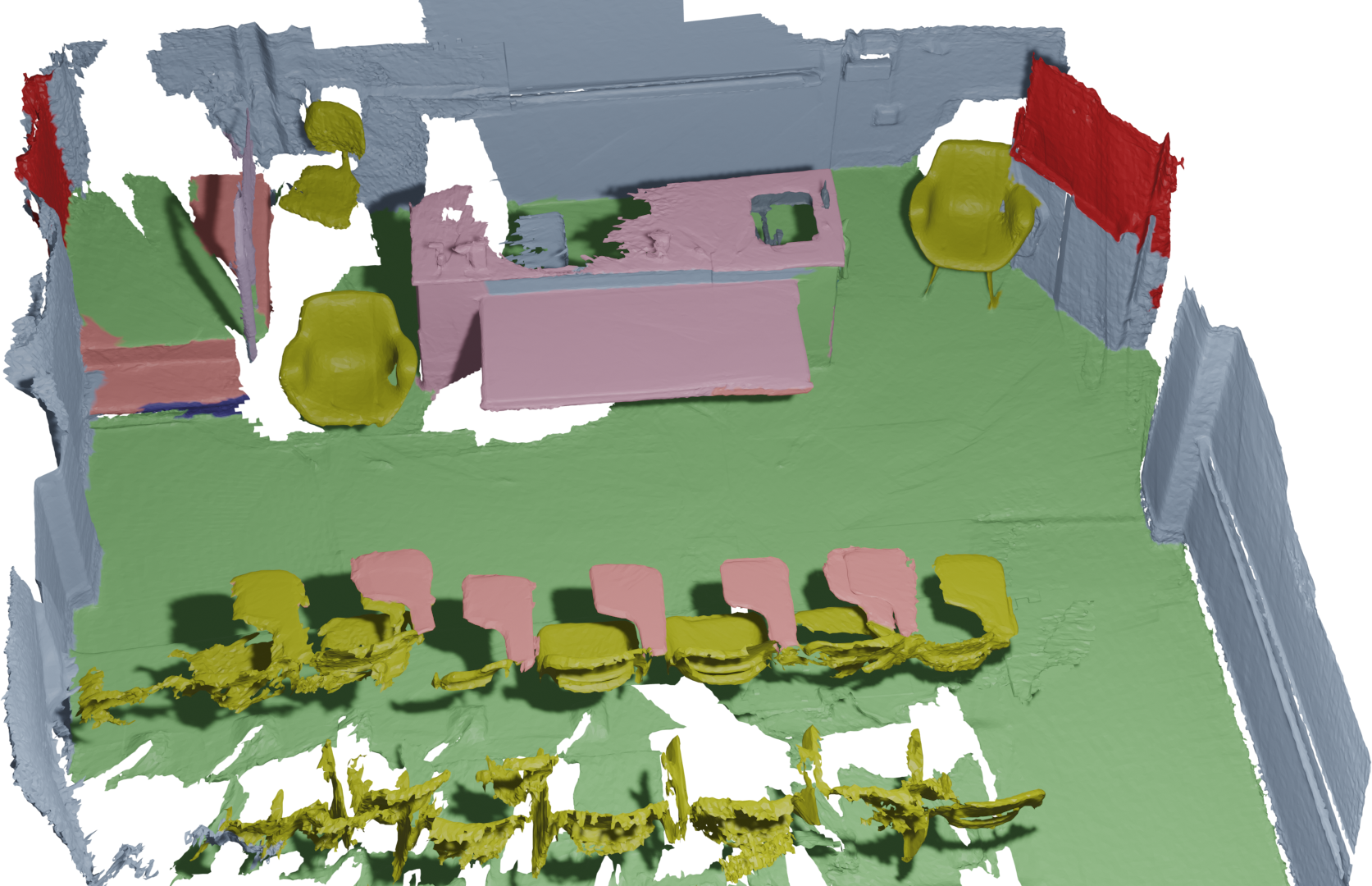}
		\caption{\footnotesize{OneFormer3D}}
		\label{fig:oneformer3d}
	\end{subfigure}
	\begin{subfigure}[b]{0.19\textwidth}
		\centering
		\includegraphics[width=\textwidth]{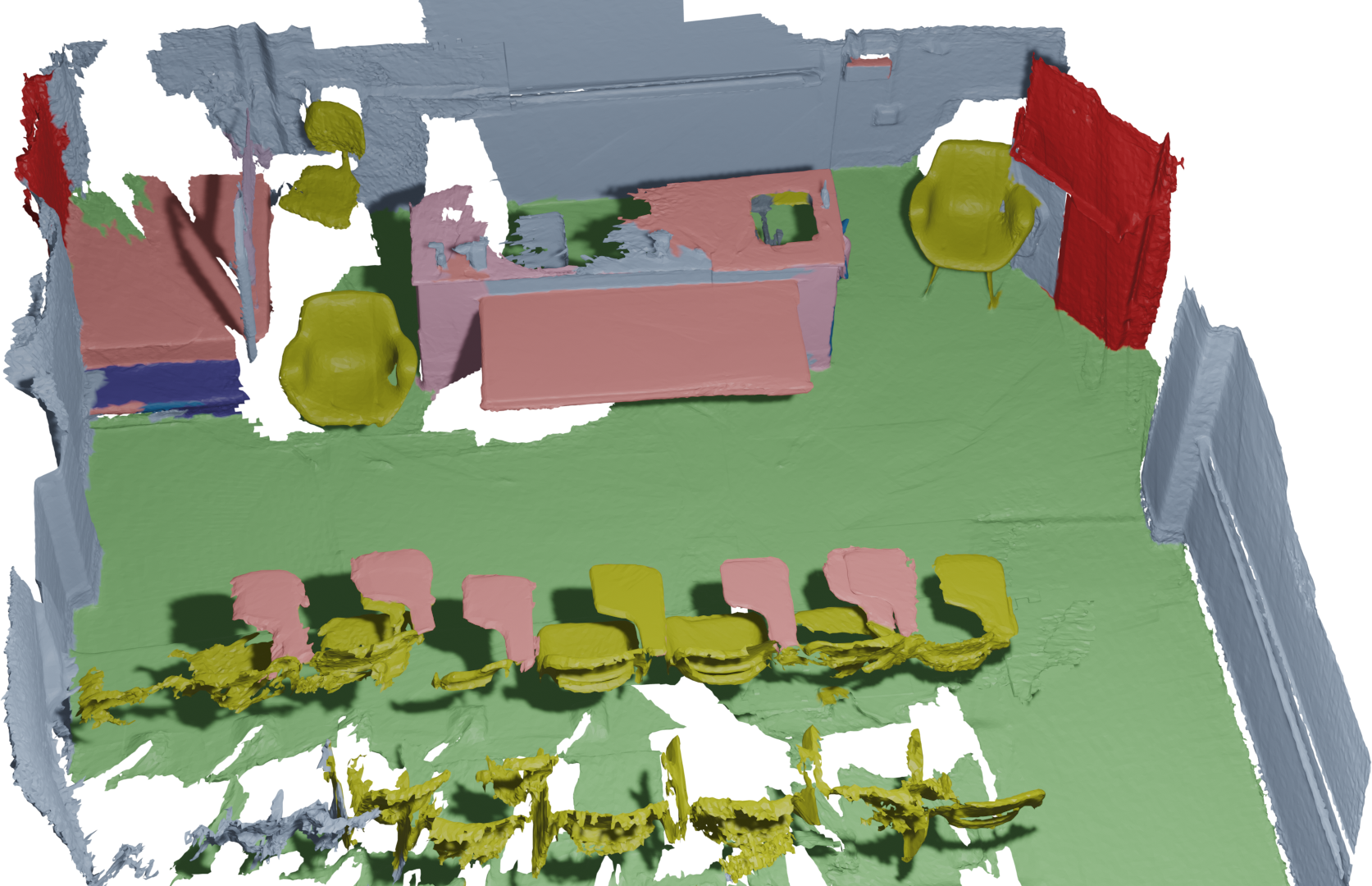}
		\caption{\footnotesize{Description Prediction}}
		\label{fig:des_pred}
	\end{subfigure}
	\begin{subfigure}[b]{0.19\textwidth}
		\centering
		\includegraphics[width=\textwidth]{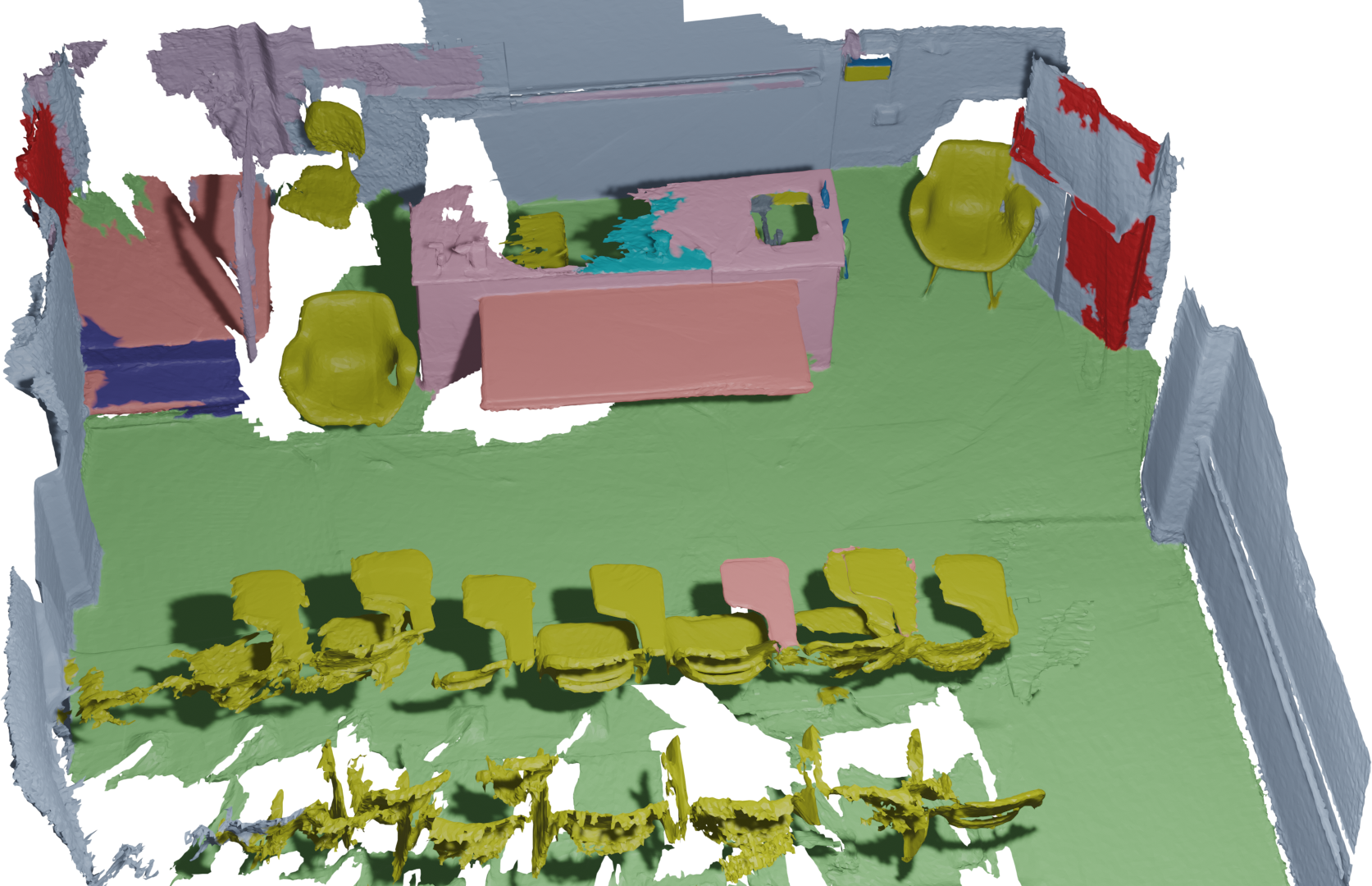}
		\caption{\footnotesize{Image Prediction}}
		\label{fig:image_pred}
	\end{subfigure}
	\begin{subfigure}[b]{0.19\textwidth}
		\centering
		\includegraphics[width=\textwidth]{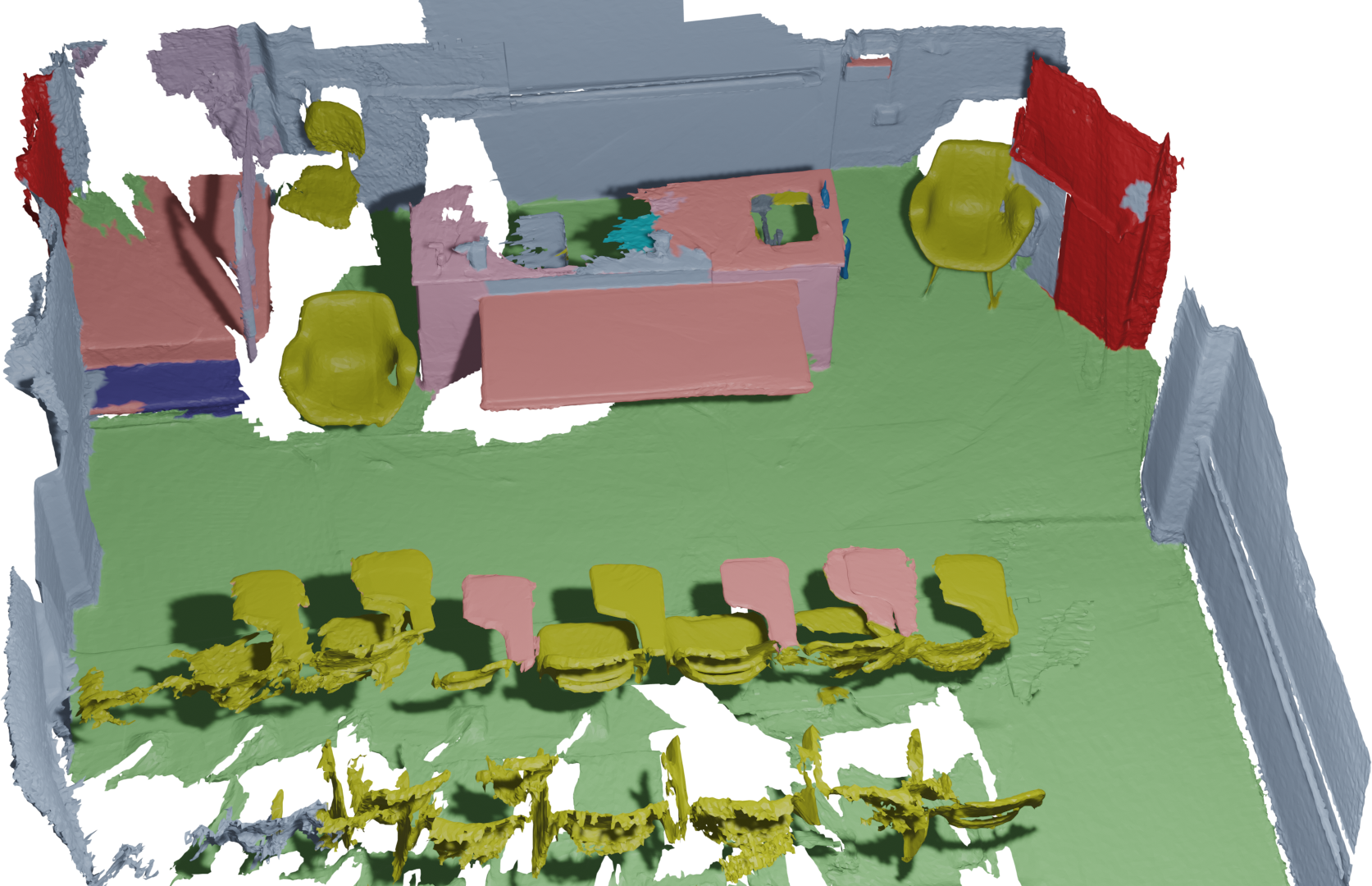}
		\caption{\footnotesize{VDG-Uni3DSeg }}
		\label{fig:all_pred}
	\end{subfigure}
    \vspace{-0.2cm}
	\caption{Qualitative results of semantic segmentation on ScanNet. Compared to OneFormer3D, the Description and Image Prediction modules improve segmentation accuracy, particularly for tables. By integrating both description and image queries, VDG-Uni3DSeg achieves more balanced and consistent segmentation, enhancing the semantic segmentation for small and challenging objects such as doors and chairs. }
	\vspace{-0.2cm}
	\label{fig:visual}
\end{figure*}

\subsection{Ablation Study}
\vspace{-0.15cm}
Table~\ref{tab:ablation} presents an ablation study evaluating the contribution of each component in our VDG-Uni3DSeg framework on S3DIS.
First, removing the Spatial Enhancement Module slightly degrades the performance, particularly in instance and panoptic segmentation metrics (e.g., a 2-point decrease in mAP and a 1.7-point drop in PQ).
This underscores the importance of spatial information in effectively distinguishing instances within the scene.
Next, we analyze the impact of reference queries.
Removing either the Description Query or the Image Query leads to a significant performance drop, with a more pronounced decline when the Description Query is removed.
This demonstrates the benefit of detailed descriptions generated by LLMs in our method.
Furthermore, incorporating reference images helps bridge the gap between textual descriptions and the 3D point cloud.
By integrating description and image queries, our method enables each point to associate with the most relevant class representation, leading to overall improvements across all segmentation tasks.
Finally, removing the SVC Loss causes the most significant drop in instance and panoptic segmentation metrics, including a 2.6-point decline in mAP\textsubscript{25} and a 6.9-point drop in PQ\textsubscript{th}.
This demonstrates that SVC Loss encourage the model to learn more discriminative segmentation features by leveraging multi-modal queries as anchors for feature alignment.
In general, each component of our method contributes to improving feature alignment and overall segmentation performance.

\subsection{Qualitative Results}
Fig.~\ref{fig:visual} illustrates the semantic segmentation results of different methods.
Compared to OneFormer3D~\cite{kolodiazhnyi2024oneformer3d}, both the Description Prediction and Image Prediction modules improve segmentation accuracy by leveraging reference queries, with notable improvements in identifying tables.
However, these individual modules struggle with segmenting complex objects such as chairs and doors.
When combining description and image predictions, as in VDG-Uni3DSeg, the segmentation results exhibit more balanced and comprehensive improvements across all object categories.
This integration results in more precise and consistent segmentation, demonstrating that while individual queries contribute to performance gains, their combined use in VDG-Uni3DSeg yields the most accurate and robust results. 
Other results are provided in the supplementary material.
\vspace{-0.1cm}

\section{Conclusion}
We present VDG-Uni3DSeg, a unified 3D segmentation framework that integrates LLM-generated descriptions and internet-sourced images as reference cues to enhance segmentation performance. 
By leveraging multimodal information, our method improves class and instance differentiation without requiring human annotations or paired multimodal data.  
To further refine segmentation, we introduce the Semantic-Visual Contrastive (SVC) Loss, which aligns point features with multimodal queries to enhance feature distinctiveness. 
Additionally, our Spatial Enhancement Module effectively captures long-range spatial relationships, ensuring precise segmentation boundaries while balancing computational cost.  
Extensive experiments on multiple datasets demonstrate that VDG-Uni3DSeg achieves state-of-the-art performance across semantic, instance, and panoptic segmentation tasks. 
These results highlight the effectiveness of multimodal guidance in 3D segmentation and pave the way for future research in leveraging external knowledge to further improve 3D scene understanding.

\section{Acknowledgment}
We would like to express our sincere gratitude to all the reviewers for their valuable and constructive suggestions, which have significantly improved the quality of this work.

{
    \small
    \bibliographystyle{ieeenat_fullname}
    \bibliography{main}
}

\clearpage
\appendix
\section*{Appendix}

\section{Ablation Study}
\textbf{About Spatial Enhancement Module (SEM)} Full attention over all points is prohibitively expensive, especially for large-scale point clouds. To address this, our SEM adopts a fixed-size subset sampling strategy, which significantly reduces computational cost while still allowing each point to access contextual information.
	We present results on ScanNet using different sampling sizes in Table~\ref{table:r3}.
	Increasing the size leads to improvements. But larger sizes will cause increased computational cost. These results, together with the ablation study, validate that the module is both effective and necessary.

\begin{table}[ht]
	\centering
	\renewcommand{\thetable}{R\arabic{table}}
	\caption{Effect of different sample size on ScanNet.}
	\centering
	\resizebox{0.95\linewidth}{!}
    {
		\label{table:r3}
		\begin{tabular}{clcccccc}
			\hline
			size& mAP\textsubscript{50} & mAP & mPrec\textsubscript{50} & mRec\textsubscript{50} & mIoU & PQ  \\
			\midrule
			16&77.0&56.9&85.3&76.5&75.0&70.1\\
			64&77.0&57.9&\textbf{86.0}&76.1&75.9&70.8\\
			128 & \textbf{78.5}& \textbf{59.3}&{85.7}&\textbf{78.3}& \textbf{76.2}& \textbf{71.5}\\ 
			\hline
		\end{tabular}
        }
\end{table}

\begin{table}[t]
	\centering
		\renewcommand{\thetable}{R\arabic{table}}
		\caption{Comparision with random images on S3DIS.}
		\centering
		\resizebox{0.9\linewidth}{!}
		{
			\label{table:r2}
			\begin{tabular}{llcccccc}
				\hline
				Method& mAP\textsubscript{50} & mAP & mPrec\textsubscript{50} & mRec\textsubscript{50} & mIoU & PQ\\
				\midrule
				rand1&72.5&58.5&79.3&72.9&\textbf{71.7}&62.0\\
				rand2&72.1&57.6&\textbf{81.9}&\textbf{75.2}&71.3&63.5\\
				Ours  & \textbf{74.1} & \textbf{60.1} & {81.0} & {73.9} & {71.5} & \textbf{66.3} \\
				
				\hline
			\end{tabular}
		}
\end{table}

\begin{figure*}[ht]
    \hsize=\textwidth 
    \centering
    \includegraphics[width=0.9\textwidth]{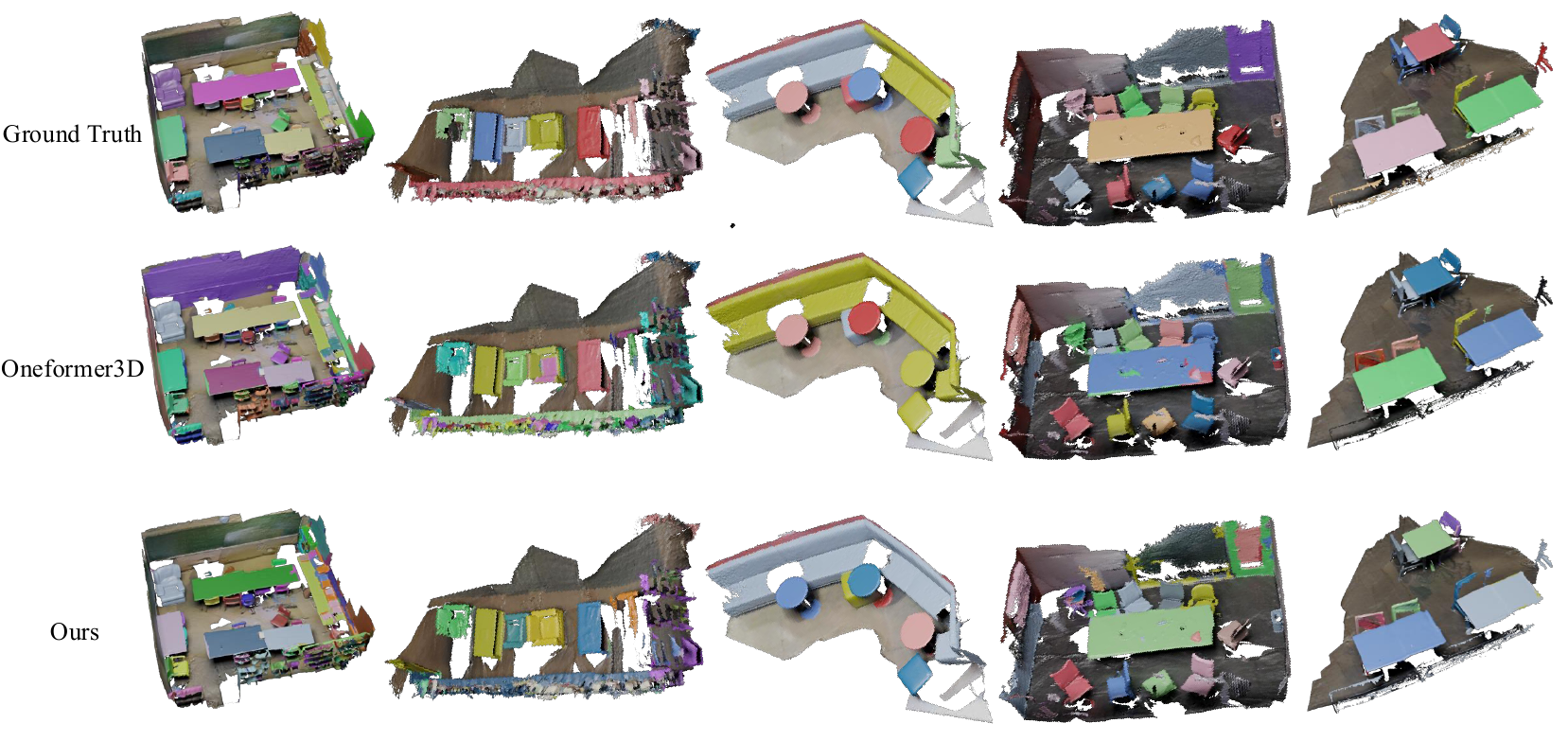}
    \vspace{-0.2cm}
        \caption{The qualitative results of the instance segmentation on ScanNet. }
        \vspace{-0.2cm}
        \label{fig:inst_visual}
\end{figure*}

\begin{figure*}[ht]
		\hsize=\textwidth 
		\centering
		\includegraphics[width=0.9\textwidth]{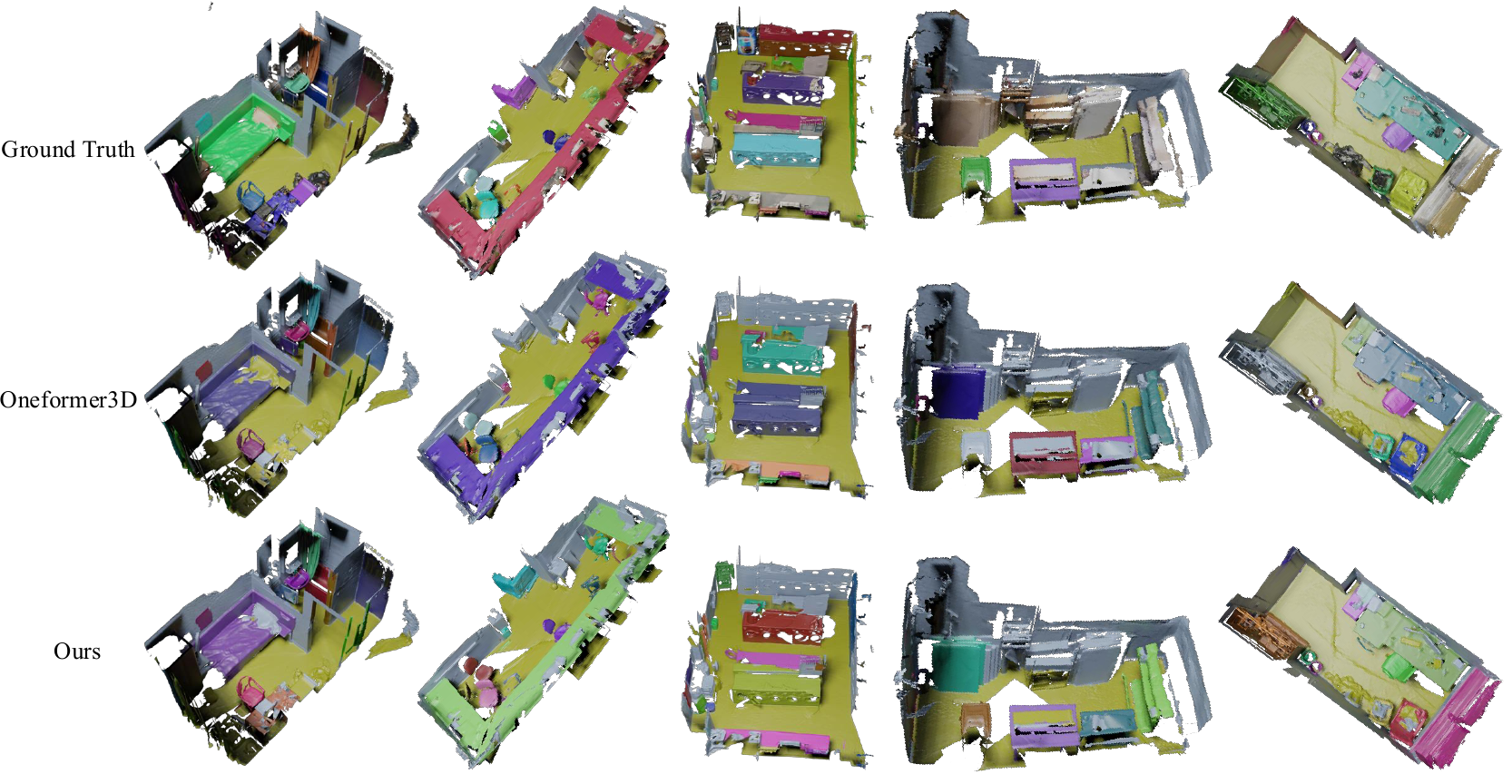}
        \vspace{-0.2cm}
		\caption{The qualitative results of the panoptic segmentation on ScanNet. }
        \vspace{-0.2cm}
		\label{fig:pano_visual}
\end{figure*}

\noindent\textbf{About internet images} 
We collect 20 images per class and select the top-5 most relevant ones based on their CLIP similarity with corresponding class name. 
We also test with 2 random image sets on S3DIS. As shown in Table~\ref{table:r2}, CLIP-selected images perform better overall, which shows that high-quality images are more beneificial, especially those paired multi-view images.
The internet images offers a lightweight and easily accessible alternative, in contrast to methods~\cite{kundu2020virtual,robert2022learning,peng2023openscene} that use paired multi-view images.
Our approach is the first to leverage a small number of unpaired general images and textual descriptions for enhancing 3D point cloud segmentation.
We agree, however, that bridging the remaining gap to models that leverage paired multi-view inputs remains a challenging problem. In future work, we plan to explore stronger image–point-cloud alignment techniques, enabling more effective use of easily collected general images to enhance 3D scene understanding, including in open-vocabulary settings.

\section{Qualitative Results}
Fig.\ref{fig:inst_visual} presents qualitative comparisons of instance segmentation results on ScanNet, where VDG-Uni3DSeg is evaluated against the baseline OneFormer3D\cite{kolodiazhnyi2024oneformer3d}.
Our approach exhibits a strong ability to distinguish individual object instances, particularly excelling in cluttered environments with complex spatial arrangements.
By leveraging LLM-generated descriptions and reference images, our method enhances class-specific feature learning, resulting in more consistent instance segmentation.

Fig.~\ref{fig:pano_visual} further showcases the panoptic segmentation results on ScanNet.
VDG-Uni3DSeg produces more coherent and consistent segmentation across both stuff and thing categories.
Notably, our approach significantly enhances the consistency of stuff regions, effectively reducing misclassification in large homogeneous areas while preserving precise instance differentiation.
These qualitative results show the robustness of our method in complex 3D scenes.

\end{document}